\pdfoutput=1

\documentclass[11pt,a4paper]{article}
\usepackage{times,latexsym}
\usepackage{url}
\usepackage[T1]{fontenc}
\usepackage{algorithm}
\usepackage{algorithmic}
\usepackage{amsmath}
\usepackage{amssymb}
\usepackage{amstext}
\usepackage{mathtools}
\usepackage{resizegather}
\usepackage{booktabs}
\usepackage[pdftex]{graphicx}
\usepackage{multirow}
\usepackage{makecell}
\usepackage[usenames,dvipsnames,table,xcdraw]{xcolor} 
\usepackage{paralist}
\usepackage{svg}

\usepackage{tikz}
\usetikzlibrary{positioning}
\usetikzlibrary{bayesnet}
\usetikzlibrary{arrows}
\usetikzlibrary{backgrounds}
\usetikzlibrary{quotes}
\usepackage{caption}
\usepackage{subcaption}
\usepackage{pgfplots} 
\usepackage{microtype}
\usepackage{svg}
\pgfplotsset{compat=1.14}
\usepackage{xspace,mfirstuc,tabulary}
\usepackage[skip=1pt]{parskip}
\newcommand{\indep}{\perp\!\!\!\perp}
\DeclareMathOperator{\Tr}{Tr}

\usepackage[acceptedWithA]{tacl2021v1}
\usepackage[capitalise]{cleveref}
\crefformat{footnote}{#2\footnotemark[#1]#3}
\AtBeginDocument{} \AtBeginDocument{%
 \abovedisplayskip=1pt plus1pt minus1pt%
 \abovedisplayshortskip=1pt plus1pt \belowdisplayskip=\abovedisplayskip
 \belowdisplayshortskip=1pt plus1pt minus1pt%
 }

\title{Optimal Transport Posterior Alignment for Cross-lingual Semantic Parsing}

\author{Tom Sherborne \and Tom Hosking \and Mirella Lapata\\
Institute for Language, Cognition and Computation\\
  School of Informatics, University of Edinburgh \\
10 Crichton  Street, Edinburgh EH8 9AB\\
  \texttt{\{tom.sherborne,~tom.hosking\}@ed.ac.uk,~mlap@inf.ed.ac.uk} \\
}

\date{}

\begin{document}
\maketitle
\begin{abstract}
Cross-lingual semantic parsing transfers parsing capability from a
high-resource language (e.g., English) to low-resource languages with
scarce training data. Previous work has primarily considered
silver-standard data augmentation or zero-shot methods, however,
exploiting few-shot gold data is comparatively unexplored. We propose
a new approach to cross-lingual semantic parsing by explicitly
minimizing cross-lingual divergence between probabilistic latent
variables using Optimal Transport. We demonstrate how this direct
guidance improves parsing from natural languages using fewer examples
and less training.  We evaluate our method on two datasets, MTOP and
MultiATIS++SQL, establishing state-of-the-art results under a few-shot
cross-lingual regime. Ablation studies further reveal that our method
improves performance even without parallel input translations. In
addition, we show that our model better captures cross-lingual
structure in the latent space to improve semantic representation
similarity.
\footnote{Our code and data are publicly available
at \href{https://github.com/tomsherborne/minotaur}{\tt
  github.com/tomsherborne/minotaur}.}

\end{abstract}


\section{Introduction}
\label{sec:introduction}

Semantic parsing maps natural language utterances to logical form (LF)
representations of meaning. As an interface between human- and computer-readable
languages, semantic parsers are a critical component in various natural language
understanding (NLU) pipelines, including assistant technologies
\citep{kollar-etal-2018-alexa}, knowledge base question answering
\citep{berant-etal-2013-semantic,Liang:2016}, and code generation
\citep{wang-etal-2023-mconala}.

Recent advances in semantic parsing have led to improved reasoning over
challenging questions \citep{li2023resdsql} and accurate generation of complex
queries \citep{scholak-etal-2021-picard}, however, most prior work has focused
on English \citep{kamath2019a,Qin2022ASO}. Expanding, or \emph{localizing}, an
English-trained model to additional languages is challenging for several
reasons. There is typically little labeled data in the target languages due to
high annotation costs. Cross-lingual parsers must also be sensitive to
how different languages refer to entities or model abstract and mathematical
relationships \citep{reddy-etal-2017-universal,hershcovich-etal-2019-semeval}.
Transfer between dissimilar languages can also degrade in multilingual models
with insufficient capacity \citep{pfeiffer-etal-2022-lifting}.

Previous strategies for resource-efficient localization include generating
``silver-standard'' training data through machine-translation
\citep{nicosia-etal-2021-translate-fill} or prompting large language models
\citep{rosenbaum-etal-2022-clasp}.
Alternatively, zero-shot models use
``gold-standard'' external corpora for auxiliary tasks \citep{van-der-goot-etal-2021-masked} and
few-shot models maximize sample-efficiency using meta-learning
\citep{10.1162/tacl_a_00533}. We argue that previous work encourages
cross-lingual transfer through \emph{implicit} alignment only via minimizing
silver-standard data perplexity, multi-task ensembling, or constraining gradients. 

We instead propose to localize an encoder-decoder semantic parser by
\emph{explicitly} inducing cross-lingual alignment between representations. We
present \mbox{\textsc{Minotaur}} (\textbf{Min}imizing \textbf{O}ptimal
\textbf{T}ransport distance for \textbf{A}lignment \textbf{U}nder
\textbf{R}epresentations)---a method for cross-lingual semantic parsing which
explicitly minimizes distances between probabilistic latent variables to reduce
representation divergence across languages (\Cref{fig:minotaur_diagram}).
{\sc Minotaur} leverages Optimal Transport theory \citep{Villani2008OptimalTO}
to measure and minimize this divergence between English and target languages
during episodic few-shot learning. Our hypothesis is that explicit alignment
between latent variables can improve knowledge transfer between languages
without requiring additional annotations or lexical alignment.  We evaluate this
hypothesis in a \emph{few-shot} cross-lingual regime and study how many examples
in languages beyond English are needed for ``good'' performance.

Our technique allows us to precisely measure, and minimize, the cross-lingual
transfer gap between languages. This yields both sample-efficient training and
establishes leading performance for few-shot cross-lingual transfer on two
datasets. We focus our evaluation on semantic parsing but {\sc Minotaur} can be
applied directly to a wide range of other tasks.  Our contributions are as
follows:


\begin{figure}
    \centering
    \includegraphics[width=\linewidth]{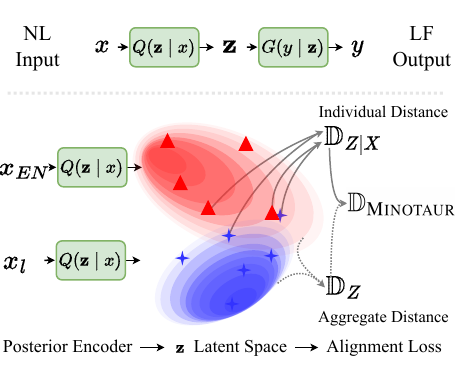}
    \caption{\textbf{Upper}: We align representations explicitly in the latent representation space, $\mathbf{z}$, between encoder $Q$ and decoder $G$. \textbf{Lower}: {\sc Minotaur} induces cross-lingual similarity by minimizing
    divergence between latent distributions at
    two levels---between individual and aggregate posteriors.
    \label{fig:minotaur_diagram}}
    \vspace*{-1.5em}
\end{figure}

\begin{itemize}
    \item We propose a method for learning a semantic parser using
    \emph{explicit} cross-lingual alignment between probabilistic latent
    variables. \mbox{\textsc{Minotaur}} jointly minimizes marginal and
    conditional posterior divergence for \emph{fast} and \emph{sample-efficient}
    cross-lingual transfer.\\
    \item We propose an episodic training scheme for cross-lingual posterior alignment during training which requires minimal modifications to typical learning.\\
    \item Experiments on task-oriented semantic parsing (MTOP;
      \citealt{li-etal-2021-mtop}) and executable semantic parsing
      (MultiATIS++SQL; \citealt{sherborne-lapata-2022-zero}) demonstrate that
      \mbox{\textsc{Minotaur}} outperforms prior methods with fewer data
      resources and faster convergence.
\end{itemize}


\section{Related Work}
\label{sec:related_work}

\paragraph{Cross-lingual Semantic Parsing} 

Growing interest in cross-lingual NLU has motivated the expansion of
benchmarks to study model adaptation across many languages
\citep{hu2020xtreme, liang-etal-2020-xglue}. Within executable
semantic parsing, ATIS \citep{hemphill-etal-1990-atis} has been
translated into multiple languages such as Chinese and Indonesian
\citep{susanto-lu-2017-neural}, and GeoQuery
\citep{geoquery-Zelle:1996:LPD:1864519.1864543} has been translated
into German, Greek, and Thai
\citep{jones-etal-2012-semantic}. Adjacent research in Task-Oriented
Spoken Language Understanding (SLU) has given rise to datasets such as
MTOP in five languages \citep{li-etal-2021-mtop}, and MultiATIS++ in
seven languages \citep{xu-etal-2020-end}. SLU aims to parse
inputs into functional representations of dialog acts (which are often
embedded in an assistant NLU pipeline) instead of executable
machine-readable language.

In all cases, cross-lingual semantic parsing demands fine-grained
semantic understanding for successful transfer across languages.
Multilingual pre-training \citep{pires-etal-2019-multilingual} has the
potential to unlock certain understanding capabilities but is often
insufficient. Previous methods resort to expensive dataset translation
\citep{jie-lu-2014-multilingual,neural-hybrid-trees-susanto2017} or
attempt to mitigate data paucity by creating ``silver'' standard data
through machine translation
\citep{sherborne-etal-2020-bootstrapping,nicosia-etal-2021-translate-fill,xia-monti-2021-multilingual,guo-etal-2021-learning}
or prompting \citep{rosenbaum-etal-2022-clasp,shi-etal-2022-xricl}.
However, methods that rely on synthetic data creation are yet to
produce cross-lingual parsing equitable to using gold-standard
professional translation.

Zero-shot methods bypass the need for \mbox{in-domain} data augmentation using
multi-task objectives which incorporate gold-standard data for external tasks
such as language modeling or dependency parsing
\citep{van-der-goot-etal-2021-masked,sherborne-lapata-2022-zero,
gritta-etal-2022-crossaligner}. Few-shot approaches which leverage a small
number of annotations have shown promise in various tasks \citep[\emph{inter
alia.}]{zhao-etal-2021-closer} including semantic parsing.
\citet{10.1162/tacl_a_00533} propose a first-order meta-learning algorithm to
train a semantic parser capable of sample-efficient cross-lingual transfer. 

Our work is most similar to recent studies on cross-lingual alignment for
classification tasks \citep{wu-dredze-2020-explicit} and spoken-language
understanding using token- and slot-level annotations between parallel inputs
\citep{qin-etal-2022-gl,liang-etal-2022-label}. While similar in motivation, we
contrast in our exploration of latent variables with parametric alignment for a
closed-form solution to cross-lingual transfer. Additionally, our method does
not require fine-grained word and phrase alignment annotations, instead inducing
alignment in the continuous latent space.

\paragraph{Alignment and Optimal Transport}

Optimal Transport (OT; \citealt{Villani2008OptimalTO}) minimizes the
cost of mapping from one distribution (e.g., utterances) to another
(e.g., logical forms) through some joint distribution with conditional
independence \cite{monge1781memoire}, i.e., a latent variable
conditional on samples from one input domain. OT in NLP has mainly
used Sinkhorn distances to measure the divergence between non-parametric
discrete distributions as an online minimization sub-problem
\citep{NIPS2013_af21d0c9}.

Cross-lingual approaches to OT have been proposed for embedding alignment
\citep{alvarez-melis-jaakkola-2018-gromov, alqahtani-etal-2021-using-optimal},
bilingual lexicon induction \citep{marchisio-etal-2022-bilingual} and
summarization \citep{Nguyen_Luu_2022}. Our method is similar to recent proposals
for cross-lingual retrieval using variational or OT-oriented representation
alignment \citep{10.1145/3539597.3570468, wieting2023contrastive}.
\citet{wang-wang-2019-riemannian} consider a ``continuous’’ perspective on OT
using the Wasserstein Auto-Encoder \citep[{\sc Wae}]{tolstikhin2018wasserstein}
as a language model which respects geometric input characteristics within the
latent space. 

Our parametric formulation allows this continuous approach to OT,
similar to the {\sc Wae} model. While monolingual prior work in
semantic parsing has identified that latent structure can benefit the
semantic parsing task
\citep{kocisky-etal-2016-semantic,yin-etal-2018-structvae}, it does
not consider whether it can inform transfer between languages. To the
best of our knowledge, we are the first to consider the continuous
form of OT for cross-lingual transfer in a sequence-to-sequence
task. We formulate the parsing task as a transportation problem in
\Cref{sec:background} and describe how this framework gives rise to
explicit cross-lingual alignment in \Cref{sec:minotaur}.


\section{Background}
\label{sec:background}
\subsection{Cross-lingual Semantic Parsing}
\label{sec:xlingual_structpred}

Given a natural language utterance $x$, represented as a sequence of tokens
$\left(x_{1},\ldots, x_{T}\right)$, a semantic parser generates a faithful
logical-form meaning representation $y$.\footnote{Notation key: Capitals~$X$,
are random variables; Curly~$\mathcal{X}$, are functional domains; lowercase~$x$
are observations and $P_{\{\}}$~are probability distributions.} A typical neural
network parser trains on input-output pairs $\lbrace
x_{i},~y_{i}\rbrace_{i=0}^{N}$, using the cross-entropy between predicted
$\hat{y}$, and gold-standard logical form $y$, as supervision
\citep{cheng-etal-2019-learning}.

Following the standard VAE framework
\citep{https://doi.org/10.48550/arxiv.1312.6114,pmlr-v32-rezende14}, an encoder
$Q_{\phi}$ represents inputs from $\mathcal{X}$ as a continuous latent variable
$Z$, $Q_{\phi}:\mathcal{X}\rightarrow\mathcal{Z}$. A decoder $G_{\theta}$
predicts outputs conditioned on samples from the latent space,
$G_{\theta}:\mathcal{Z}\rightarrow\mathcal{Y}$. The encoder therefore acts as
approximate posterior~$Q_{\phi}(Z|X)$. $Q_{\phi}$ is a multi-lingual pre-trained
encoder shared across all languages.

For cross-lingual transfer, the parser must also generalize to
languages from which it has seen few (or zero) training
examples.\footnote{Resource parity between languages is
  \emph{multilingual} semantic parsing which we view as an
  upper-bound.} Our goal is for the prediction for input $x_{l} \in
X_{l}$ in language~$l$ to match the prediction for equivalent input
from a high-resource language (typically English), i.e.,~\mbox{$x_{l}
  \rightarrow y,~x_{\rm EN} \rightarrow y$} subject to the constraint
of fewer training examples in $l$ ($\lvert N_{l}\rvert\ll~\lvert
N_{\rm EN}\rvert$). As shown in \Cref{fig:minotaur_diagram}, we
propose measuring the divergence between approximate posteriors
(i.e.,~$Q\left(Z|X_{\rm EN}\right)$ and $Q\left(Z|X_{l}\right)$) as
the distance between individual samples and an approximation of the
``mean'' encoding of each language. This goal of aligning
distributions naturally fits an Optimal Transport perspective.


\subsection{Kantorovich Transportation Problem}
\label{sec:wasserstein}

\setlength{\belowdisplayskip}{10pt} \setlength{\belowdisplayshortskip}{10pt}
\setlength{\abovedisplayskip}{10pt} \setlength{\abovedisplayshortskip}{10pt}

\citet{tolstikhin2018wasserstein} propose the Wasserstein Auto-Encoder ({\sc
Wae}) as an alternative variational model. The {\sc Wae} minimizes the
\emph{transportation cost} under the Kantorovich form of the Optimal Transport
problem \citep{KantorovichOT}. Given two distributions $P_X, P_Y$, the objective
is to find a \emph{transportation plan} ~$\Gamma\left(X,~Y\right)$, within the
set of all joint distributions, $\mathcal{P}\left(X\sim P_{X},~Y\sim
P_{Y}\right)$, to map probability mass from $P_{X}$ to $P_{Y}$ with minimal
cost. $T_{c}$~expresses the problem of finding a plan which minimizes a
transportation cost function
$c\left(X,~Y\right):~\mathcal{X}\times\mathcal{Y}\rightarrow\mathcal{R}_{+}$:
\begin{equation}
    \begin{split}
        T_{c}\left(P_{X}, P_{Y}\right) &\coloneqq \\
        \inf_{\scriptscriptstyle\Gamma\in\left(X\sim P_{X},~Y\sim P_{Y}\right)}&\mathbb{E} _{\left(X,Y\right)\sim\Gamma}\left[c\left(X,~Y\right)\right]
        \label{eq:ot_kantorovich}
    \end{split}    
\end{equation}
The {\sc Wae} is proposed as an auto-encoder (i.e.,~$P_{Y}$ approximates
$P_{X}$), however, in our setting $P_{X}$ is the natural language input
distribution and $P_{Y}$ is the logical form output distribution and they
are both realizations of the same semantics.

Using conditional independence, ${y}\indep{x}~|~\textbf{z}$, we can transform
the plan, $\Gamma\left(X,~Y\right)\rightarrow\Gamma\left(Y|X\right)P_{X}$ and
consider a non-deterministic mapping from $X$ to $Y$ under observed $P_{X}$.
\citet[Theorem 1]{tolstikhin2018wasserstein} identify how to \emph{factor} this
mapping through latent variable $Z$, leading to:
\[
\begin{split}
        T_{c}\left(P_{X}, P_{Y}\right) = \hspace{-0.75em}\inf_{\scriptscriptstyle Q_{\phi}\left(Z|X\right)\in\mathcal{Q}}\hspace{-0.5em}&\mathbb{E}_{P_{X}}\mathbb{E}_{Q_{\phi}\left(Z|X\right)} \left[c\left(Y, G_{\theta}\left(Z\right)\right)\right] \\ 
        &+ \alpha \mathbb{D}\left(Q(Z), P(Z)\right)
    \end{split}    
\]\vspace{-1.36cm}
\begin{equation}
\label{eq:ot_wae_objective}
\end{equation}
\Cref{eq:ot_wae_objective}  expresses a minimizable objective:~identify the
probabilistic encoder~$Q_{\phi}\left(Z|X\right)$ and decoder
$G_{\theta}\left(Z\right)$ which minimizes a cost, subject to regularization on
the divergence~$\mathbb{D}$ between the marginal posterior~$Q\left(Z\right)$ and
prior~$P\left(Z\right)$. 

The additional regularization is how the {\sc Wae} improves on the evidence
lower bound in the variational auto-encoder, where the equivalent alignment on
the individual posterior~$Q_{\phi}(Z|X)$ drives latent representations to zero.
Regularization on the marginal posterior
\mbox{$Q\left(Z\right)=\mathbb{E}_{X\sim
P_{X}}\left[Q_{\phi}\left(Z|X\right)\right]$} instead allows individual
posteriors for different samples to remain distinct and non-zero. This limits
posterior collapse, guiding~$Z$ to remain informative for decoding.

We use Maximum Mean Discrepancy \citep[\sc Mmd]{JMLR:v13:gretton12a} for an
unbiased estimate of $\mathbb{D}\left(Q(Z), P(Z)\right)$ as a robust measure of
the distance between high dimensional Gaussian distributions.
\Cref{eq:mmd_k_rkhs} defines {\sc Mmd} using some kernel
$k:~\mathcal{Z}\times\mathcal{Z}\ \rightarrow\mathcal{R} $, defined over a
reproducible kernel Hilbert space, $\mathcal{H}_{k}$:
\begin{equation}
  \begin{array}{ll}
        \textsc{Mmd}_{k}\left({\scriptstyle P,~Q}\right) = & \\
&       \hspace*{-1.8cm} \lVert \int_{\mathcal{Z}}{\displaystyle k\left(z,\cdot\right)
          dP}{\scriptstyle~-}\int_{\mathcal{Z}}{\displaystyle
          k\left(z,\cdot\right)
          dQ}\rVert_{\mathcal{H}_k}
\end{array}
\label{eq:mmd_k_rkhs}
\end{equation}
Informally, {\sc Mmd} minimizes the distance between the ``feature means'' of
variables $P$ and $Q$ estimated over a batch sample.  \Cref{eq:mmd_ustatistic}
defines {\sc Mmd} estimation over observed $\mathbf{p}$ and $\mathbf{q}$ using
the heavy-tailed \emph{inverse multiquadratic} ({\sc Imq}) kernel~$k$:
\[
\begin{split}
        &\hspace*{-.6cm}\textsc{Mmd}_{k}\left(\mathbf{p},\mathbf{q}\right) =
  {\displaystyle \frac{1}{n_{p}\left(n_{p}-1\right)}}\sum_{z'\neq
    z}k(p_{z},~p_{z'})+
\end{split}
\]\vspace*{-1.8cm}
\begin{equation}
  \label{eq:mmd_ustatistic}
\end{equation}
\vspace*{-.5cm}
\begin{equation}
  \begin{split}
        &{\displaystyle \frac{1}{n_{q}\left(n_{q}-1\right)}}\sum_{z'\neq z}k(q_{z},~q_{z'}) - 
        {\displaystyle \frac{2}{n_{p}n_{q}}}\sum_{z,~z'}k(p_{z},~q_{z'}) \nonumber
        \end{split}
\end{equation}
We define the \textsc{Imq} kernel in \Cref{eq:mmd_imq_kernel} below;
$C=2|\mathbf{z}|\sigma^2$  and $\mathcal{S}=\left[0.1,0.2,0.5,1,2,5,10 \right]$.
\begin{equation}
\label{eq:mmd_imq_kernel}
  k\left(p, q\right)=\sum_{s\in \mathcal{S}}\frac{s\cdot C}{s\cdot C + \lVert p-q \rVert_{2}^{2}}
\end{equation}

This framework defines a {\sc Wae} objective using a cost function, $c$ to map from $P_{X}$ to $P_{Y}$ through latent variable $Z$. We now describe how {\sc Minotaur} integrates explicit posterior alignment during this learning process.

\section{{\sc Minotaur}:~Posterior Alignment for Cross-lingual Transfer}
\label{sec:minotaur}
\paragraph{Variational Encoder-Decoder} 

Our model comprises of encoder (and approximate posterior)~$Q_{\phi}$, and
generator decoder~$G_{\theta}$.  The encoder~$Q_{\phi}$ produces a distribution
over latent encodings \mbox{$\mathbf{z}=\{z_{1},\ldots,~z_{T}\}$}, parameterized
as a sequence of $T$ mean states $\boldsymbol{\mu}_{\{1,\ldots,T\}}\in
\mathbb{R}^{T \times d}$, and a single variance $\sigma^{2}\in \mathbb{R}^{d}$
for all $T$ states,
\begin{align} 
\mathbf{z} = Q_{\phi}\left(x\right) \sim \mathcal{N} (\boldsymbol{\mu},~\sigma^{2}).
    \label{eq:encoder_1}
\end{align}
The latent encodings $\mathbf{z}$ are sampled using the Gaussian
reparameterization trick \citep{https://doi.org/10.48550/arxiv.1312.6114},
\begin{align}
    \textbf{z}&=\boldsymbol{\mu}+\sigma^{2}\circ\boldsymbol{\epsilon},~\boldsymbol{\epsilon}\sim
    \mathcal{N}(\textbf{0},\,\textbf{I}).
    \label{eq:encoder_2}
\end{align}
Finally, an output sequence $\hat{y}$ is generated from $\mathbf{z}$ through
autoregressive generation,
\begin{align}
\hat{y}&=G_{\theta}\left(\mathbf{z}\right) \label{eq:decoder_1}
\end{align}

For an input sequence of $T$ tokens, we use a sequence of $T$ latent variables
for $\mathbf{z}$ over pooling into a single representation. This allows for more
`bandwidth' in the latent state to minimize the risk of the decoder ignoring
$\mathbf{z}$ i.e.,~\emph{posterior collapse}. We find this design choice to be
necessary as lossy pooling leads to weak overall performance. We also
use a single variance estimate for sequence $\mathbf{z}$---this minimizes
variance noise across $\mathbf{z}$ and simplifies computation in posterior
alignment. We follow the convention of an isotropic unit Gaussian prior,
$P\left(\textbf{z}\right)\sim\mathcal{N}(\textbf{0}, \textbf{I})$. 

\paragraph*{Cross-lingual Alignment}

Typical {\sc Wae} modeling builds meaningful latent structure by aligning the 
estimated posterior to the prior only. {\sc Minotaur} extends this through additionally
aligning posteriors \emph{between} languages. Consider learning the optimal mapping from English utterances~$X_{\rm
  EN}$ to logical forms~$Y$ within \Cref{eq:ot_kantorovich} via latent
variable~$Z$, from monolingual data $\left( X_{\rm EN}, Y\right)$. The
optimization in \Cref{eq:ot_wae_objective} converges on an optimal
transportation plan $\Gamma_{\rm EN}^{\ast}$ as the minimum
cost.\footnote{$\Gamma_{\ast}$ is implicit within the model
  parameters.}

For transfer from English to language $l$, previous work either requires token
alignment between $X_{\rm EN}$ and $X_{l}$ or exploits the shared $Y$ between
$X_{\rm EN}$ and $X_{l}$ \citep[\it inter alia.]{qin-etal-2022-gl}. We instead
induce alignment by explicitly matching $Z$ between languages. Since $Y$ is
dependent only on $Z$, the latent variable offers a continuous representation
space for alignment with the minimal and intuitive condition that equivalent~$z$
yields equivalent~$y$. Therefore, our proposal is a straightforward extension of
learning $\Gamma_{\rm EN}^{\ast}$; we propose to bootstrap the transportation
plan for target language $l$ (i.e., $\Gamma_{l}^{\ast}\left(X_{l},~Y\right)$) by
aligning on $Z$ in a few-shot learning scenario. {\sc Minotaur}
\emph{explicitly} aligns $Z_{l}$ (from a target language $l$) towards $Z$ (from
EN) by matching $Q(Z_{l}|X_{l})$ to $Q(Z|X_{\rm EN})$ for the goal
$\Gamma_{l}^{\ast}=\Gamma_{\rm EN}^{\ast}$, thereby transferring the learned
capabilities from high-resource languages with only a few training examples.

Given parallel inputs $x_{\rm EN}$ and $x_{l}$ in English and language $l$, with
equivalent LF ($y_{\rm EN}=y_{l}$), their latent encodings are given by:
\begin{align} 
    \mathbf{z}_{\rm EN} &= Q_{\phi}\left(x_{\rm EN}\right),\hat{y}_{\rm EN}=
    G\left(\mathbf{z}_{\rm EN}\right) \label{eq:fwd_l1}\\
    \mathbf{z}_{l} &= Q_{\phi}\left(x_{l}\right),\hat{y}_{l}=
    G\left(\mathbf{z}_{l}\right). \label{eq:fwd_l2} 
\end{align}

Unlike vanilla VAEs, where $\textbf{z}$ is a single vector, the posterior
samples ($\mathbf{z}_{\rm EN}, \mathbf{z}_{l} \in \mathbb{R}^{T \times d}$) are
complex structures. We therefore follow \citet{DBLP:conf/icml/MathieuRST19} in
using a decomposed alignment signal minimizing both \emph{aggregate} posterior
alignment (higher-level) and \emph{individual} posterior alignment (lower-level)
with scaling factors $\left(\alpha_{P},\beta_{P} \right)$ respectively. This
leads to the {\sc Minotaur} alignment outlined in \Cref{fig:minotaur_diagram}
and expressed below,
\begin{multline}
    \mathbb{D}_{\textsc{Minotaur}}\left(\mathbf{z}_{\rm EN},\mathbf{z}_{l}\right) = \\
    \alpha_{P}
    \mathbb{D}_{Z}\left(Q_{\phi}\left(\mathbf{z}_{\rm EN}\right),Q_{\phi}\left(\mathbf{z}_{l}\right)
    \right) \\
        + \beta_{P} \mathbb{D}_{Z|X}\left(Q_{\phi}\left(\mathbf{z}_{l}|x_{l}\right) \|
    Q_{\phi}\left(\mathbf{z}_{\rm EN}|x_{\rm EN}\right) \right). \label{eq:loss_minotaur}
\end{multline}
where~$\mathbb{D}_{Z|X}$ is a divergence penalty between \emph{individual}
representations to match local structure, while $\mathbb{D}_{Z}$ is a divergence
penalty between representation \emph{aggregates} to match more global structure.
The intuition is that individual matching promotes contextual encoding
similarity and aggregate matching promotes similarity at the language level.

Similar to the prior alignment, we use the {\sc Mmd} distance to align
aggregate posteriors as \Cref{eq:mmd_k_rkhs} (i.e., marginal posteriors over $Z$ between languages). For individual alignment, we consider two numerically
stable \emph{exact} solutions to measure individual divergence which are well
suited to matching high-dimensional Gaussians \citep{ojm/1326291215}. Modeling
$Q_{\phi}\left(Z|X\right)$ as a parametric statistic yields the benefit of
\mbox{closed-form} computation during learning. We primarily use the $L^{2}$
Wasserstein distance,~$W_{2}$, as the Optimal Transport-derived minimum
transportation cost between Gaussians ($\mathbf{p},\mathbf{q}$) across domains.
Within \Cref{eq:wasserstein_l2} the mean is~$\mu$, covariance is~
\mbox{$\Sigma={\rm Diag}\lbrace \sigma^2_{i},\ldots,\sigma^2_{n} \rbrace$}, and
encodings have dimensionality~$d$. ${\rm Tr}\lbrace\rbrace$ is the matrix trace
function. 

\begin{equation} 
  \begin{array}{l}
\hspace*{-3cm}    W_{2}\left(\mathbf{p},\mathbf{q}\right) = ||
      \mu_{\mathbf{p}} - \mu_{\mathbf{q}} ||_{2}^{2}~+ \\
    \end{array}  \label{eq:wasserstein_l2}
\end{equation}
\vspace*{-1.1cm}
\begin{equation}
  \begin{split}
        &\hspace{2cm}\Tr\{\Sigma_\mathbf{p}+\Sigma_\mathbf{q}-2\left(\Sigma_\mathbf{p}^{\frac{1}{2}}\Sigma_\mathbf{q}\Sigma_\mathbf{p}^{\frac{1}{2}}\right)^{\frac{1}{2}}
    \}. \end{split} \nonumber
\end{equation}

We also consider the Kullback-Leibler Divergence (KL) between two Gaussian
distributions as \Cref{eq:kl_gaussians}. Minimizing KL is
equivalent to maximizing the mutual information between distributions
as an information-theoretic goal of semantically aligning
$\mathbf{z}$. \Cref{sec:results} demonstrates that $W_{2}$ is superior
to KL in all cases.

\begin{equation}
    \begin{split}
        \hspace*{-.2cm}{\rm KL}&\left(\mathbf{p}\|\mathbf{q}\right) =
        \frac{1}{2}\left(\log\left(\frac{|\Sigma_\mathbf{q}|}{|\Sigma_\mathbf{p}|}\right)~-~d_{p,q}~+
        \right.\\
    \end{split}
  \label{eq:kl_gaussians}
\end{equation}
\vspace*{-.8cm}
\begin{equation}
  \begin{split}
        &\hspace{.6cm}\left.\Tr\lbrace\Sigma_\mathbf{q}^{-1}\Sigma_\mathbf{p}\rbrace+\left(\mu_{\mathbf{q}}-\mu_{\mathbf{p}}\right)^{T}\Sigma_\mathbf{q}\left(\mu_{\mathbf{q}}-\mu_{\mathbf{p}}\right) \right) 
    \end{split}.  \nonumber
  \end{equation}
${\rm
  Tr}\lbrace\rbrace$ is the matrix trace function. Minimizing KL is
equivalent to maximizing the mutual information between distributions
as an information-theoretic goal of semantically aligning
$\mathbf{z}$. \Cref{sec:results} demonstrates that $W_{2}$ is superior
to KL in all cases.

We express $\mathbb{D}_{Z|X}$ (see \Cref{eq:loss_minotaur}) between singular
$\mathbf{p}$ and $\mathbf{q}$ representations for individual tokens for clarity,
however, we actually minimize the \emph{mean} of~$\mathbb{D}_{Z|X}$ between
each~$\mathbf{z}_1$ and~$\mathbf{z}_2$ tokens across both sequences,
i.e.,~$\frac{1}{|\mathbf{z}_1||\mathbf{z}_2|}\sum_{i,j}\mathbb{D}_{Z|X}\left(\mathbf{z}_{1i}\|\mathbf{z}_{2j}\right)$.
We observe that minimizing this mean divergence between all
$\left(\mathbf{z}_{1i},\mathbf{z}_{2j}\right)$ pairs is most empirically
effective.

Finally, \Cref{eq:loss_fn} expresses the transportation cost, $T_{c}$, for a single
$\left(x,~y\right)$ pair during training: the cross-entropy between predicted
and gold $y$ and {\sc Wae} marginal prior regularization.

\begin{multline}
    \mathcal{L}{\left(x,y\right)}= \mathbb{E}_{Q\left(\mathbf{z}|x\right)}
    \left[-\sum_{i}y_{i}\left(\log G_{\theta}\left(\mathbf{z}\right)\right)_{i}\right] +\\
        \alpha\mathbb{D}\left(Q_{\phi}(\mathbf{z}),
    P(\mathbf{z})\right) \label{eq:loss_fn} 
    \end{multline}
We episodically augment \Cref{eq:loss_fn} as \Cref{eq:loss_total} using the {\sc
Minotaur} loss every $k$ steps for few-shot induction of cross-lingual
alignment. Sampling $(x, y)$ is detailed in \Cref{sec:exp_setup}.

\begin{equation}
\begin{array}{r} 
\hspace*{-1cm}\mathcal{L}_{\Sigma} = \mathcal{L}{\left(x_{\rm EN},y_{\rm
            EN}\right)} + 
        \mathcal{L}{\left(x_{l},y_{l}\right)} \\
     +~~   \mathbb{D}_{\textsc{Minotaur}}\left(\mathbf{z}_{\rm EN},\mathbf{z}_{l}\right)\\
\end{array}        \label{eq:loss_total} 
\end{equation}

Another perspective on our approach is that we are aligning pushforward
distributions, \mbox{$Q\left(X\right):\mathcal{X}\rightarrow\mathcal{Z}$}.
Cross-lingual alignment at the input token level (in $\mathcal{X}$) requires
fine-grained annotations and is an outstanding research problem (see
\Cref{sec:related_work}). Our method of aligning pushforwards in $\mathcal{Z}$
is smoothly continuous, does not require word alignment, and does not always
require input utterances to be parallel translations. While we evaluate {\sc
Minotaur} principally on semantic parsing, our framework can extend to any
sequence-to-sequence or representation learning task which may benefit from
explicit alignment between languages or domains. 


\section{Experimental Setting}
\label{sec:exp_setup}

\paragraph{MTOP \citep{li-etal-2021-mtop}} contains dialog utterances of
``assistant'' queries and their corresponding tree-structured slot and intent
LFs. MTOP is split into 15,667 training, 2,235 validation, and 4,386 test
examples in English (EN). A variable subsample of each split is translated into
French (FR), Spanish (ES), German (DE), and Hindi (HI). We refer to \citet[Table
1]{li-etal-2021-mtop} for complete dataset details. As shown in
\Cref{tab:io-examples}, we follow \citet[Appendix
B.2]{rosenbaum-etal-2022-clasp} using ``space-joined'' tokens and ``sentinel
words'' (i.e., a {\tt wordi} token is prepended to each input token and replaces
this token in the LF) to produce a closed decoder vocabulary
\cite{raman-etal-2022-transforming}. This allows the output LF to reference
input tokens by label without a copy mechanism. We evaluate LF accuracy using
the \emph{Space and Case Invariant Exact-Match} metric (SCIEM;
\citealt{rosenbaum-etal-2022-clasp}).

We sample a small number of training instances for low-resource
languages, following the \emph{Samples-per-Intent-and-Slot} (SPIS)
strategy from \citet{chen-etal-2020-low} which we adapt to our
cross-lingual scenario. SPIS randomly selects examples and keeps those
that mention any slot and intent value (e.g.,~``{\tt IN:}'' and
``{\tt SL:}'' from \Cref{tab:io-examples}) with fewer than some rate
in the existing subset.  Sampling stops when all slots and intents
have a minimum frequency of the sampling rate (or the maximum if fewer
than the sampling rate). SPIS sampling ensures a minimum coverage of
all slot and intent types during cross-lingual transfer. This
normalizes unbalanced low-resource data as the model has seen
approximately similar examples across all semantic
categories. Practically, an SPIS rate of 1, 5, 10 equates to 284
(1.8\%), 1,125 (7.2\%), and 1,867 (11.9\%) examples (\%~training
data).

\paragraph{MultiATIS++SQL \citep{sherborne-lapata-2022-zero}} Experiments on
ATIS \citep{hemphill-etal-1990-atis} study cross-lingual transfer using an
executable LF to retrieve database information. We use the MultiATIS++SQL
version (see \Cref{tab:io-examples}), pairing executable SQL with parallel
inputs in English (EN), French (FR), Portuguese (PT), Spanish (ES), German (DE),
and Chinese (ZH). We measure \emph{denotation accuracy}---the proportion of
executed predictions retrieving equivalent database results as executing the
gold LF. Data is split into 4,473 training, 493 validation, and 448 test examples
with complete translation for all splits. We follow \citet{10.1162/tacl_a_00533}
in using random sampling. Rates of 1\%, 5\%, and 10\% correspond to 45, 224, and
447 examples respectively. For both datasets, the model only observes remaining
data in English, e.g., sampling at 5\% uses 224 multilingual examples and 4,249
English-only examples for training.


{\setlength{\tabcolsep}{6pt}
\begin{figure}[t]
    \centering
    
    \begin{tabulary}{\columnwidth}{@{}LL@{}} \toprule \textcolor{ForestGreen}{$x_{\rm
    EN}$} & {\footnotesize {\tt word1} Who {\tt word2} attended {\tt
    word3} Yale?}  \\ \textcolor{ForestGreen}{$x_{\rm DE}$} &
    {\footnotesize {\tt word1} Wer {\tt word2} besuchte {\tt word3}
    Yale?}\\ \textcolor{ForestGreen}{$y$} & {\footnotesize\tt
    {[}IN:GET\_CONTACT {[}SL:SCHOOL {\tt word3} {]}{]}
    } \\ \midrule \textcolor{Red}{$x_{\rm EN}$} & {\footnotesize What
    does ORD mean?} \\ \textcolor{Red}{$x_{\rm FR}$} & {\footnotesize
    Que signifie ORD?} \\ \textcolor{Red}{$y$} & {\footnotesize\tt
    SELECT DISTINCT airport.airport\_name FROM airport WHERE
    airport.code=ORD;} \\ \bottomrule \end{tabulary} \caption{Input,
    $x$, and output, $y$, examples in English (EN), German (DE) and
    French (FR) for MTOP \citep[upper
    {\color{ForestGreen}green}]{li-etal-2021-mtop} and
    MultiATIS++SQL \citep[lower
    {\color{red}red}]{sherborne-lapata-2022-zero},
    respectively.\label{tab:io-examples}}
\end{figure}
}

\paragraph{Modeling}
We follow prior work in using a Transformer encoder-decoder: we use
the frozen pre-trained 12-layer encoder from {\sc mBART50}
\citep{tang-etal-2021-multilingual} and append an identical learnable
layer. The decoder is a six-layer Transformer stack
\citep{transformers-noam-Vaswani2017AttentionIA} matching the encoder
dimensionality ($d=1,024$). Decoder layers are trained from scratch
following prior work and early experiments verified that pre-training
the decoder did not assist in cross-lingual transfer, offering minimal
improvement on English.  The variance predictor ($\sigma^{2}$ for
predicting $\mathbf{z}$ in \Cref{eq:encoder_1}) is a multi-head pooler
from \citet{liu-lapata-2019-hierarchical} adapting multi-head
attention to produce singular output from sequential inputs. The final
model has $\sim$116 million trainable parameters and $\sim$340 million
frozen parameters.

\paragraph{Optimization} 
We train for a maximum of ten epochs with early stopping using validation loss.
Optimization uses Adam \citep{DBLP:journals/corr/KingmaB14} with a batch size of
256 and learning rate of $1\times 10^{-4}$. We empirically tune hyperparameters
$\left(\beta_{P},\alpha_{P}\right)$ to $\left(0.5, 0.01\right)$ respectively.
During learning, a typical step (without {\sc Minotaur} alignment) samples a
batch of $\left(x_{L}, y\right)$ pairs in languages $L\in\{{\rm EN}, l_{1},
l_{2}\ldots\}$ from a sampled dataset described above. Each {\sc Minotaur} step
instead uses a sampled batch of parallel data $\left(x_{\rm EN},~x_{l},~y_{\rm
EN},~y_{l}\right)$ to induce explicit cross-lingual alignment from the same data
pool. The episodic learning loop size is tuned to $k=20$; we find that if $k$ is
infrequent then posterior alignment is weaker and if $k$ is too frequent then
overall parsing degrades as the posterior alignment dominates learning. Tokenization uses SentencePiece
\citep{kudo-richardson-2018-sentencepiece} and beam search prediction uses five
hypotheses. All experiments are implemented in PyTorch \citep{pytorch} and
AllenNLP \citep{gardner-etal-2018-allennlp}. Training takes one hour using
1$\times$ A100 80GB GPU for either dataset.

\paragraph{Comparison Systems}

As an upper-bound, we train the {\sc Wae}-derived model without
low-resource constraints. We report monolingual (one language) and
multilingual (all languages) versions of training a model on available
data. We use the monolingual upper-bound EN model as a
``Translate-Test'' comparison. We also compare to monolingual and
multilingual ``Translate-Train'' models to evaluate the value of gold
samples compared to silver-standard training data. We follow previous
work in using OPUS \citep{tiedemann-2012-parallel} translations for
MTOP and Google Translate \citep{gtranslate} for MultiATIS++SQL in all
directions. Following \citet{rosenbaum-etal-2022-clasp}, we use a
cross-lingual word alignment tool (SimAlign;
\citealt{jalili-sabet-etal-2020-simalign}) to project token positions
from MTOP source to the parallel machine-translated output (e.g., to
shift label {\tt wordi} in EN to {\tt wordj} in FR).

In all results, we report averages of five runs over different
few-shot splits.  For MTOP, we compare to ``silver-standard'' methods:
``Translate-and-Fill'' \citep[{TaF}]{nicosia-etal-2021-translate-fill}
which generates training data using MT, and {CLASP}
\citep{rosenbaum-etal-2022-clasp} which uses MT and prompting to
generate multilingual training data. We note that these models and
dataset pre-processing methods are not public (we have confirmed that
our methods are reasonably comparable with authors). For
MultiATIS++SQL, we compare to {\sc XG-Reptile} from
\citep{10.1162/tacl_a_00533}. This method uses meta-learning to
approximate a ``task manifold'' using English data and constrain
representations of target languages to be close to this manifold. This
approach \emph{implicitly} optimizes for cross-lingual transfer by
regularizing the gradients for target languages to align with
gradients for English. {\sc Minotaur} differs in \emph{explicitly}
measuring the representation divergence across languages.

\section{Results}
\label{sec:results}

We find that {\sc Minotaur} validates our hypothesis that \emph{explicitly}
minimizing latent divergence improves cross-lingual transfer with few training
examples in the target language.  As evidenced by our ablation studies, our
technique is surprisingly robust and can function without any parallel data
between languages. Overall, our method outperforms silver-standard data
augmentation techniques (in \Cref{tab:main:mtop}) and few-shot meta-learning (in
\Cref{tab:main:atis}).

\paragraph*{Cross-lingual Transfer in Task-Oriented Parsing}


\begin{table*}[t]
\centering
\resizebox{\linewidth}{!}{%
\begin{tabular}{@{}lcccccc@{}}
\toprule
    & EN    & FR    & ES    & DE    & HI    & Avg. \\ \midrule
Gold Monolingual & 79.4 & 69.8 & 72.3 & 67.1 & 60.5 & 67.4$~\pm~$5.3 \\
Gold Multilingual & {\bf 81.3} & {\bf 75.7} & {\bf 77.2} & {\bf 72.8} & {\bf 71.6} & {\bf 74.4$~\pm~$3.5} \\ \midrule
Translate-Test   & ---   & 7.7  & 7.4  & 7.6  & 7.3  & 7.5$~\pm~$0.2 \\
Translate-Train Monolingual  & ---   & 41.7 & 31.4 & 50.1 & 32.2 & 38.9$~\pm~$9.4 \\
Translate-Train Multilingual & 74.2 & 46.9 & 43.0 & 53.6 & 39.9 & 45.9$~\pm~$5.9 \\ 
Translate-Train Multilingual $+${\sc Minotaur} & 77.5 & 59.9 & 60.2 & 61.6 & 42.2 & 56.0$~\pm~$9.2 \\ \midrule
TaF mT5-large \citep{nicosia-etal-2021-translate-fill}   & 83.5 & 71.1 & 69.6 & 70.5 & 58.1 & 67.3$~\pm~$6.2 \\
TaF mT5-xxl \citep{nicosia-etal-2021-translate-fill}  & {\bf 85.9} & {\bf 74.0} & 71.5 & {\bf 72.4} & 61.9 & 70.0$~\pm~$5.5      \\
CLASP \citep{rosenbaum-etal-2022-clasp} & 84.4 & 72.6 & 68.1 & 66.7 & 58.1 & 66.4$~\pm~$6.1 \\ \midrule
{\sc Minotaur}~1 SPIS  & 79.5$~\pm~$0.4 & 71.9$~\pm~$0.2 & 72.3$~\pm~$0.1 & 68.4$~\pm~$0.3 & 65.1$~\pm~$0.1 & 69.4$~\pm~$3.4 \\
{\sc Minotaur}~5 SPIS  & 77.7$~\pm~$0.6 & 72.0$~\pm~$0.6   & 73.6$~\pm~$0.3 & 69.1$~\pm~$0.5 & 68.2$~\pm~$0.5 & 70.7$~\pm~$2.5 \\
{\sc Minotaur}~10 SPIS & 80.2$~\pm~$0.4 & 72.8$~\pm~$0.5 & {\bf 74.9}$~\pm~$0.1 & 70.0$~\pm~$0.7  & {\bf 68.6}$~\pm~$0.5 & {\bf 71.6}$~\pm~$2.8 \\ \bottomrule
\end{tabular}%
}
\caption{Accuracy on MTOP across (i) upper-bounds, (ii) translation
  baselines, (iii) ``silver-standard'' methods, and (iv) {\sc Minotaur}
  with SPIS sampling at 1, 5 and 10. We report for \textit{English},
  \textit{French}, \textit{Spanish}, \textit{German}, and
  \textit{Hindi} with $\pm$ sample standard deviation. \textit{Avg.}
  reports the target language average $\pm$ standard deviation across
  languages. Best result per-language and average for (i) and
  (ii)--(iv) are bolded.  }
\label{tab:main:mtop}
\end{table*}

\Cref{tab:main:mtop} summarizes our results on MTOP against comparison
models at multiple SPIS rates. Our system significantly improves on
the ``Gold Monolingual'' upper-bound even at 1 SPIS by~\mbox{$>2\%$}
(\mbox{$p<0.01$}, using a two-tailed sign test assumed hereafter). For
few-shot transfer on MTOP, we observe strong cross-lingual transfer
even at 1 SPIS translating only 1.8\% of the dataset. Few-shot
transfer is competitive with a monolingual model using 100\% of gold
translated data and so represents a promising new strategy for this
dataset. We note that even at a high SPIS rate of 100
(approximately~$\sim53.1$\% of training data), {\sc Minotaur} is
significantly (\mbox{$p<0.01$}) poorer than the ``Gold Multilingual''
upper-bound, highlighting that few-shot transfer is challenging on
MTOP.

{\sc Minotaur} outperforms all translation-based comparisons; augmenting
``Translate-Train Multilingual'' with our posterior alignment objective
($+$~\textsc{Minotaur}) yields a~$+10.1\%$ average improvement. With equivalent
data, this comparison shows that cross-lingual alignment by aligning
each latent representation to the prior \emph{only} (i.e., a {\sc Wae}-based
model) is weaker than cross-lingual alignment between posteriors.

\paragraph{Comparing to ``Silver-Standard'' Methods}
A more realistic comparison is between {TaF}
\cite{nicosia-etal-2021-translate-fill} or {CLASP}
  \cite{rosenbaum-etal-2022-clasp}, which optimize MT quality in their
  pipelines, and our method which uses sampled gold data. We
  outperform CLASP by $>$3\% and TaF using mT5-large
  \citep{xue-etal-2021-mt5} by $>$2.1\% at \emph{all} sample
  rates. However, {\sc Minotaur} requires $>5$ SPIS sampling to
  improve upon TaF using mT5-xxl. We highlight that our model has only
  $\sim116$ million parameters whereas CLASP uses AlexaTM-500M
  \citep{FitzGerald2022} with 500 million parameters, mT5-large has
  700 million parameters and mT5-xxl has 3.3 billion
  parameters. Relative to model size, our approach offers improved
  computational efficiency. The improvement of our method is mostly
  seen in languages typologically distant from English as {\sc
    Minotaur} is always the strongest model for Hindi. In contrast,
  our method underperforms for English and German (more similar to EN)
  which may benefit from stronger pre-trained knowledge transfer
  within larger models. Our efficacy using gold data and a smaller
  model, compared to silver data in larger models, suggests a quality
  trade-off, constrained by computation, as a future study.

\paragraph*{Cross-lingual Transfer in Executable Parsing}


\begin{table*}[!ht]
\centering
\resizebox{\textwidth}{!}{%
\begin{tabular}{@{}llccccccc@{}}
\toprule
\multicolumn{2}{l}{} & EN & FR & PT & ES  & DE  & ZH  & Avg. \\ \midrule
\multicolumn{2}{l}{Gold Monolingual} & 72.3 & 73.0 & 71.8 & 67.2  & 73.4  & {\bf 73.7}  & 71.9$~\pm~$2.7  \\
\multicolumn{2}{l}{Gold Multilingual}    & {\bf 73.7} & {\bf 74.4} & {\bf 72.3} & {\bf 71.7} & {\bf 74.6}  & 71.3  & {\bf 72.9}$~\pm~$1.5  \\ \midrule
\multicolumn{2}{l}{Translate-Test}   & ---    & 70.1 & 70.6 & 66.9  & 68.5  & 62.9  & 67.8$~\pm~$3.1  \\
\multicolumn{2}{l}{Translate-Train Monolingual}  & ---    & 62.2 & 53.0 & 65.9  & 55.4  & 67.1  & 60.8$~\pm~$6.3  \\
\multicolumn{2}{l}{Translate-Train Multilingual} & 72.7 & 69.4 & 67.3 & 66.2  & 65.0  & 69.2  & 67.5$~\pm~$1.9  \\
\multicolumn{2}{l}{Translate-Train Multilingual $+${\sc Minotaur}} & 74.8 & 73.7 & 71.3 & 68.5  & 70.1  & 69.0  & 70.6$~\pm~$2.1  \\ \midrule
\multirow{2}{*}{@1\%} & \mbox{\sc XG-Reptile} & 73.8$~\pm~$0.3 & 70.4$~\pm~$1.8 & 70.8$~\pm~$0.7 & 68.9$~\pm~$2.3 & 69.1$~\pm~$1.2 & 68.1$~\pm~$1.2 & 69.5$~\pm~$1.1 \\
 & {\sc Minotaur}  & {\bf 75.6}$~\pm~$0.4 & {\bf 73.7}$~\pm~$0.6 & {\bf 71.4}$~\pm~$0.9 & {\bf 71.0}$~\pm~$0.5   & {\bf 70.4}$~\pm~$1.3 & {\bf 70.0}$~\pm~$0.9   & {\bf 71.3}$~\pm~$1.4 \\ \midrule
\multirow{2}{*}{@5\%}    & \mbox{\sc XG-Reptile} & 74.4$~\pm~$1.3 &
73.0$~\pm~$0.9 & 71.6$~\pm~$1.1 & {\bf 71.6}$~\pm~$0.7 & 71.1$~\pm~$0.6 &
69.5$~\pm~$0.5 & 71.4$~\pm~$1.3 \\
 & {\sc Minotaur}  & {\bf 77.0}$~\pm~$1.0 & {\bf 73.9}$~\pm~$1.4 & {\bf 72.8}$~\pm~$1.1 & 71.1$~\pm~$0.6 & {\bf 72.8}$~\pm~$2.0   & {\bf 72.3}$~\pm~$0.6 & {\bf 72.6}$~\pm~$1.0 \\ \midrule
\multirow{2}{*}{@10\%}   & \mbox{\sc XG-Reptile} & 75.8$~\pm~$1.3 &
74.2$~\pm~$0.2 & 72.8$~\pm~$0.6 & 72.1$~\pm~$0.7 & 73.0$~\pm~$0.6 & {\bf
72.8}$~\pm~$0.5 & 73.0$~\pm~$0.8 \\
 & {\sc Minotaur}  & {\bf 79.8}$~\pm~$0.4 & {\bf 75.6}$~\pm~$1.8 & {\bf 75.4}$~\pm~$0.8 & {\bf 73.2}$~\pm~$1.7 & {\bf 76.8}$~\pm~$1.5 & 72.5$~\pm~$0.7 & {\bf 74.7}$~\pm~$1.8 \\ \bottomrule 
\end{tabular}%
}
\caption{Denotation Accuracy on MultiATIS++SQL across (i) upper-bounds, (ii) translation baselines, and (iii) few-shot sampling for {\sc Minotaur} compared to {\sc XG-Reptile} \citep{10.1162/tacl_a_00533} at 1\%, 5\%, and 10\%. We report for
\emph{English},~\emph{French},~\emph{Portuguese},~\emph{Spanish},~
\emph{German}, and \emph{Chinese} $\pm$ sample standard deviation.
\emph{Avg.} reports the target language average $\pm$ standard deviation across
languages. Best result per-language and average for (i) and (ii)--(iii) are bolded.}
\label{tab:main:atis}
\end{table*}

The results for MultiATIS++SQL in \Cref{tab:main:atis} show similar
trends. However, here {\sc Minotaur} can outperform the upper-bounds,
and sampling at~\mbox{$>5$\%} significantly (\mbox{$p<0.01$}) improves
on ``Gold-Monolingual'' and is similar or better than
``Gold-Multilingual'' (\mbox{$p<0.05$}). Further increasing the sample
rate yields marginal gains.
{\sc Minotaur} generally improves on {\sc XG-Reptile} and performs on par
\emph{at a lower sample rate}, i.e.~{\sc Minotaur} at 1\% sampling is closer to
{\sc XG-Reptile} at 5\% sampling. This suggests that our approach is \emph{more
sample efficient}, achieving greater accuracy with fewer samples. {\sc Minotaur}
requires $<10$ epochs to train whereas {\sc XG-Reptile} reports $\sim50$
training epochs, for poorer results.

Despite demonstrating overall improvement, {\sc Minotaur} is not universally
superior. Notably, our performance on Chinese (ZH) is weaker than {\sc
XG-Reptile} at 10\% sampling and our method appears to benefit less from more
data in comparison. The divergence minimization in {\sc Minotaur} may be more
functionally related to language similarity (dissimilar languages demanding
greater distances to minimize) whereas the alignment via gradient constraints
within meta-learning could be less sensitive to this phenomenon. These results,
with the observation that {\sc Minotaur} improves most on Hindi for MTOP,
illustrate a need for more in-depth studies of cross-lingual transfer between
\emph{distant} and \emph{lower resource} languages. Future work can consider
more challenging benchmarks across a wider pool of languages
\citep{ruder2023xtremeup}.



\paragraph*{Contrasting Alignment Signals}

We report ablations of {\sc Minotaur} on MTOP at 10 SPIS sampling.
\Cref{tab:mtop:ablation:align} considers each function for
cross-lingual alignment outlined in \Cref{sec:wasserstein} as an
individual or composite element. The best approach, used in all other
reported results, minimizes the Wasserstein distance $W_{2}$ for
\emph{individual} divergence and MMD for \emph{aggregate}
divergence. $W_{2}$ is significantly superior to the Kullback-Leibler
Divergence (KL) for minimizing \emph{individual} posterior samples
(\mbox{$p<0.01$} for individual and joint cases). The $W_{2}$ distance
directly minimizes the Euclidean~$L_{2}$ distance when variances of
different languages are equivalent. This in turn is more similar to
the Maximum Mean Discrepancy function (the best singular objective)
which minimizes the distance between approximate ``means'' of each
distribution i.e., between $Z$ marginal distributions. Note that MMD
and $W_{2}$ alignments are not significantly different ($p=0.08$). The
$W_{2}$ + MMD approach significantly outperforms all other
combinations (\mbox{$p<0.01$}). The identified strength of MMD,
compared to methods for computing $\mathbb{D}_{Z|X}$, highlights that
minimizing \emph{aggregate} divergence is the main contributor to
alignment with \emph{individual} divergence as a weaker additional
contribution.

\begin{table}[t]
\centering
\resizebox{\columnwidth}{!}{%
\begin{tabular}{@{}llcccccc@{}}
\toprule
$\mathbb{D}_{Z|X}$ & $\mathbb{D}_{Z}$ & EN   & FR   & ES   & DE   & HI   & Avg. \\ \midrule
KL                 & ---   & 78.3 & 70.6 & 73.1 & 67.0 & 66.6 & 69.3 \\
$W_{2}$       & ---   & 78.6 & 72.1 & 74.3 & 68.7 & 67.4 & 70.6 \\
---                & MMD   & 78.7 & 72.3 & 74.3 & 68.8 & 67.5 & 70.7 \\
KL                 & MMD   & 78.4 & 71.8 & 73.3 & 68.5 & 67.3 & 70.2 \\
$W_{2}$       & MMD   & 80.2 & 72.8 & 74.9 & 70.0 & 68.6 & {\bf 71.6}  \\ \bottomrule
\end{tabular}%
}
\caption{Accuracy on MTOP at 10 SPIS permuting different alignment
  methods between individual-only ($\mathbb{D}_{Z|X}$), aggregate-only
  ($\mathbb{D}_{Z}$) and joint ($\mathbb{D}_{Z|X} +
  \mathbb{D}_{Z}$). The joint method using \mbox{$L_{2}$-Wasserstein}
  distance is empirically optimal but not
  significantly above the aggregate-only method ($p=0.07$).
}\label{tab:mtop:ablation:align}
\end{table}

\paragraph*{Alignment without Latent Variables}

\Cref{tab:ablation:mtop:nonparametricbneck} considers alignment
without the latent variable formulation on an encoder-decoder
Transformer model
\citep{transformers-noam-Vaswani2017AttentionIA}. Here, the output of
the encoder is not probabilistically bound without the parametric
``guidance'' of the Gaussian reparameterization. This is similar to
analysis on explicit alignment from \citet{wu-dredze-2020-explicit}.
We test MMD, statistical KL divergence (e.g., $\sum_{x}
p(x)log\left(\frac{p(x)}{q(x)}\right)$) and Euclidean~$L_{2}$ distance
as minimization functions and observe all techniques are significantly
weaker (\mbox{$p<0.01$}) than counterparts outlined in
\Cref{tab:mtop:ablation:align}. This contrast suggests the smooth
curvature and bounded structure of the $Z$ parameterization
contribute to effective cross-lingual alignment. Practically, these
non-parametric approaches are challenging to implement. The lack of
precise divergences (i.e.,~\Cref{eq:kl_gaussians} or
\Cref{eq:wasserstein_l2}) between representations leads to numerical
underflow instability during training. This impeded alignment against
typically reasonable comparisons such as cosine distance. Even using
MMD, which does not require an exact solution, fared poorer without
the bounding of the latent variable $Z$.

\begin{table}[t]
\centering
\resizebox{\columnwidth}{!}{%
\begin{tabular}{@{}lcccccc@{}}
\toprule
&  EN   & FR   & ES   & DE   & HI   &  Avg. \\ \midrule
MMD     & 77.5 & 69.6 & 70.7 & 66.3 & 61.7 & 67.1   \\
KL      & 77.9 & 69.8 & 70.9 & 66.5 & 62.1   & 67.3   \\
$L_{2}$      & 77.1 & 69.2   & 70.3 & 65.8 & 61.7 & 66.8 \\ \bottomrule
\end{tabular}%
}

\caption{Accuracy on MTOP at 10 SPIS using non-parametric alignment
  without $Z$. Here the encoder output, $E_{\phi}\left(X\right)$ is
  input into decoder
  $G_{\theta}\left(E_{\phi}\left(X\right)\right)$. All approaches
  significantly underperform (\mbox{$p<0.01$}) relative to
  \Cref{tab:mtop:ablation:align}.}
\label{tab:ablation:mtop:nonparametricbneck}
\end{table}

\paragraph*{Parallelism in Alignment}

We further investigate whether {\sc Minotaur} induces cross-lingual transfer
when aligning posterior samples from inputs which are \emph{not} parallel (i.e.,
$x_{l}$ is not a translation of $x_{\rm EN}$ and output LFs are not equivalent).
We intuitively expect parallelism as necessary for the model to minimize
divergence between representations with equivalent semantics.

As shown in \Cref{tab:ablation:mtop:nonparallel}, data parallelism is
surprisingly \emph{not required} using MMD to align marginal
distributions \emph{only}. The $\mathbb{D}_{Z|X}$ only and
$\mathbb{D}_{Z|X}+\mathbb{D}_{Z}$ techniques significantly
under-perform relative to equivalent methods using parallel data
(\mbox{$p<0.01$}). This is largely expected because individual
alignment between posterior samples which \emph{should likely not be
  equivalent} could inject unnecessary noise into the learning
process.  However, MMD ($\mathbb{D}_{Z}$ only) is significantly
(\mbox{$p<0.01$}) above other methods with the closest performance to
the parallel equivalent. This supports our interpretation that MMD
aligns ``at the language level'' as minimization between languages
should not mandate parallel data. For lower-resource scenarios, this
approach could over-sample less data for cross-lingual transfer to the
long tail of under-resourced languages.



\begin{table}[t]
\centering
\resizebox{\columnwidth}{!}{%
\begin{tabular}{@{}lcccccc@{}}
\toprule
Alignment                         & EN   & FR   & ES   & DE   & HI   & Avg. \\ \midrule
Parallel Ref. & 80.2 & 72.8 & 74.9 & 70.0 & 68.6 & {\bf 71.6} \\ 
\midrule
$\mathbb{D}_{Z|X}$ only & 78.9 & 67.3 & 68.3   & 64.6 & 59.4 & 64.9   \\
$\mathbb{D}_{Z}$ only   & 77.6 & 71.5   & 72.9 & 68.4 & 67.2 & {\bf 70.0}   \\
$\mathbb{D}_{Z|X}+\mathbb{D}_{Z}$ & 78.8 & 70.9 & 71.9 & 67.9 & 64.5 & 68.8 \\
\bottomrule
\end{tabular}%
}
\caption{Accuracy on MTOP at 10 SPIS using non-parallel inputs between
languages in {\sc Minotaur}. During training, we sample English input,~$x_{\rm EN}$, and an input in language $l$, $x_{l}$ which is \emph{not} a translation of $x_{\rm EN}$ for \Cref{eq:loss_total}. This approach weakens
individual posterior alignment but identifies that MMD is the least sensitive to input parallelism.}

\label{tab:ablation:mtop:nonparallel}
\end{table}

\paragraph*{Learning a Latent Semantic Structure}

We study the representation space learned from our method training on
MultiATIS++SQL at 1\% sampling for direct comparison to similar
analysis from \citet{10.1162/tacl_a_00533}. We compute sentence
representations from the test set as the average of the $\mathbf{z}$
representations for each input utterance
$\left(\frac{1}{T}\sum_{i}^{T}z_{i}\right)$. \Cref{tab:analysis:ranking_cosine}
compares between {\sc Minotaur}, {\sc mBART50}
\citep{tang-etal-2021-multilingual} representations before training,
and {\sc XG-Reptile}. The significant improvement in cross-lingual
cosine similarity using {\sc Minotaur} in
\Cref{tab:analysis:ranking_cosine} (\mbox{$p<0.01$}) further supports
how our proposed method learns improved cross-lingual similarity.

\begin{figure}[t]
    \centering
    \includegraphics[width=\columnwidth]{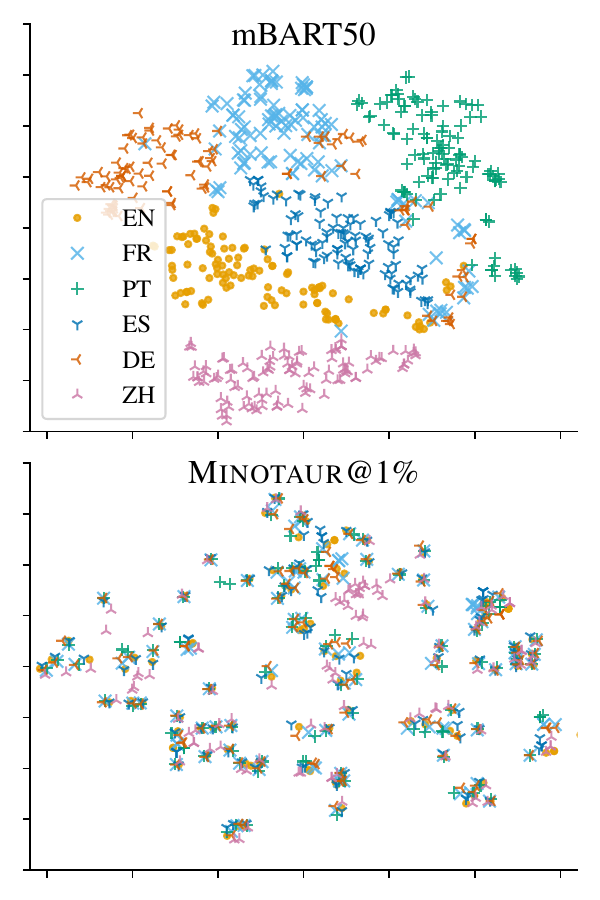}
    \caption{Visualization of MultiATIS++SQL encodings (test set; 25\%
      random parallel sample) using \mbox{t-SNE}
      \citep{JMLR:v9:vandermaaten08a}. Compared to {\sc mBART50}, {\sc
        Minotaur} organizes the latent space to be more
      \emph{semantically distributed} across languages without monolingual
      separability.}
    \label{fig:minotaur_tsne}
\end{figure}

We also consider the most cosine-similar neighbors for each representation and
test if the top-$k$ closest representations are from a parallel utterance in a
\emph{different} language or some other utterance in the \emph{same} language.
\Cref{tab:analysis:ranking_cosine} shows that $>$~99\% of representations
learned by {\sc Minotaur} have a parallel utterance within five closest
representations and $\sim$50\% improvement in mean-reciprocal ranking score
(MRR) between parallel utterances. We interpret this as the representation space
using {\sc Minotaur} is more \emph{semantically distributed} relative to {\sc
mBART50}, as representations for a given utterance are closer to semantic
equivalents. We visualize this in \Cref{fig:minotaur_tsne}:~the original
pre-trained model has minimal cross-lingual overlap, whereas our system produces
encodings with similarity aligned by \emph{semantics} rather than
\emph{language}.  {\sc Minotaur} can rapidly adapt the pre-trained
representations using an explicit alignment objective to produce a non-trivial informative latent structure. This formulation could have further utility within
multilingual representation learning or information retrieval, e.g.,~to induce
more coherent relationships between cross-lingual semantics.

\begin{table}[t]
\centering
\resizebox{\columnwidth}{!}{%
\begin{tabular}{@{}lc|cccc@{}}
\toprule
Model         & Cosine $\left(\uparrow\right)$ & Top-1 & Top-5 & Top-10 & MRR $\left(\uparrow\right)$ \\ \midrule
{\sc mBART50} & 0.576                         & 0.521 & 0.745 & 0.796   & 0.622                       \\
{\sc XG-Reptile} & 0.844 & 0.797 & 0.949 & 0.963 & 0.865  \\
{\sc Minotaur}  & {\bf 0.941}                         & {\bf 0.874} & {\bf 0.994} & {\bf 0.998}   & {\bf 0.927}                       \\ \bottomrule
\end{tabular}%
}
\caption{Average similarity between encodings of English and target
  languages for MultiATIS++SQL. Cosine similarity evaluates average
  distance between encodings of parallel sentences. Top-$k$ evaluates
  if the parallel encoding is ranked within the~$k$ most
  cosine-similar vectors. Mean Reciprocal Rank (MRR) evaluates average
  position of parallel encodings ranked by similarity. Significant best results are bolded ($p<0.01$)} 
\label{tab:analysis:ranking_cosine}
\end{table}

\paragraph*{Error Analysis}

We conduct an error analysis on MultiATIS++SQL examples correctly predicted by
{\sc Minotaur} and incorrectly predicted by baselines. The primary improvement
arises from improved handling of multi-word expressions and language-specific
modifiers. For example, adjectives in English are often multi-word adjectival
phrases in French (e.g., ``cheapest'' $\rightarrow$ ``le moins cher'' or
``earliest'' $\rightarrow$ ``\`a plus tot''). Improved handling of this error type
accounts for an average of 53\% of improvement across languages with the highest in
French (69\%) and lowest in Chinese (38\%). We hypothesize that a combination of
aggregate and mean-pool individual alignment in {\sc Minotaur} benefits this
specific case where semantics are expressed in varying numbers of words between
languages. While this could be similarly approached using fine-grained token
alignment labels, {\sc Minotaur} improves transfer in this context without
additional annotation. While this analysis is straightforward for French, it is
unclear why the transfer to Chinese is weaker. A potential interpretation is
that weaker transfer of multi-word expressions to Chinese could be related to
poor tokenization. Sub-optimal sub-word tokenization of logographic or
information-dense languages is an ongoing debate
\citep{hofmann-etal-2022-embarrassingly, 10.1162/tacl_a_00560} and exact
explanations require further study. Translation-based models and weaker systems
often generate malformed, non-executable SQL. Most additional improvement is due
to a 23\%~boost in generating syntactically well-formed SQL evaluated within a
database. Syntactic correctness is critical when a parser encounters a rare
entity or unfamiliar linguistic construction and highlights how our model can
better navigate inputs from languages minimally observed during training. This
could potentially be further improved using recent incremental decoding
advancements \citep{scholak-etal-2021-picard}.


\section{Conclusion}
\label{sec:conclusion}

We propose {\sc Minotaur}, a method for few-shot cross-lingual semantic parsing
leveraging Optimal Transport for knowledge transfer between languages. {\sc
Minotaur} uses a multi-level posterior alignment signal to enable sample-efficient
semantic parsing of languages with few annotated examples. We identify how {\sc
Minotaur} aligns individual and aggregate representations to bootstrap parsing
capability from English to multiple target languages. Our method is robust to
different choices of alignment metrics and does not mandate parallel data for
effective cross-lingual transfer. In addition, {\sc Minotaur} learns more
semantically distributed and language-agnostic latent representations with
verifiably improved semantic similarity, indicating its potential application to
improve cross-lingual generalization in a wide range of other tasks.

\section*{Acknowledgements}
We thank the action editor and anonymous reviewers for their
constructive feedback. The authors also thank Nikita Moghe, Mattia
Opper, and N. Siddarth for their insightful comments on earlier
versions of this paper. The authors (Sherborne, Lapata) gratefully
acknowledge the support of the UK Engineering and Physical Sciences
Research Council (grant EP/W002876/1). This work was supported in part
by the UKRI Centre for Doctoral Training in Natural Language
Processing, funded by the UKRI (grant EP/S022481/1) and the University
of Edinburgh (Hosking).

\bibliography{anthology,auxiliary}
\bibliographystyle{acl_natbib}

\end{document}